# Reliability of Large Language Model Generated Clinical Reasoning in Assisted Reproductive Technology: Blinded Comparative Evaluation Study

Dou Liu [1,2,6*], Ying Long [1,6,7*], Sophia Zuoqiu [3], Di Liu [3,4,5], Kang Li [4,5], Yiting Lin [8], Hanyi Liu [8], Rong Yin [3,4,5**], Tian Tang [1,6,7**]

[1] Department of Obstetrics and Gynecology, West China Second University Hospital, Sichuan University, Chengdu, China
[2] Department of Industrial and Operations Engineering, University of Michigan, Ann Arbor, US
[3] Department of Industrial Engineering, Sichuan University, Chengdu, China
[4] West China Biomedical Big Data Center, West China Hospital, Sichuan University, Chengdu, China
[5] Med-X Center for Informatics, Sichuan University, Chengdu, China
[6] Key Laboratory of Birth Defects and Related Diseases of Women and Children (Sichuan University), Chengdu, China
[7] Reproductive Medical Center, Department of Obstetrics and Gynecology, West China Second University Hospital, Sichuan University, Chengdu, China
[8] West China School of Medicine, Sichuan University, Chengdu, China
*These authors contributed equally to this work.
**These authors jointly supervised this work

**Corresponding Author:**

Tian Tang, MD

Reproductive Medical Center

Department of Obstetrics and Gynecology

West China Second University Hospital

Sichuan University

No.37, Guoxue Lane, Wuhou District

Chengdu 610041, China

Phone: +86 18081110066

Email: tiantang2016@scu.edu.cn

**Funding:** Part of this study was supported by the Science and Technology Department of Sichuan Province Project (2024YFFK0365); Part of this study was supported by the Natural Science Foundation of Sichuan, China (2025NSFSC1985); Part of this study was supported by the 1·3·5 project for disciplines of excellence, West China Hospital, Sichuan University (ZYYC21004).

# Abstract

**Background:** Creating high-quality clinical Chains-of-Thought (CoTs) is crucial for explainable medical Artificial Intelligence (AI) while constrained by data scarcity. Although Large Language Models (LLMs) can synthesize medical data, their clinical reliability remains unverified.

**Objective:** This study evaluates the reliability of LLM-generated CoTs and investigates prompting strategies to enhance their quality.

**Methods:** In a blinded comparative study, senior clinicians in Assisted Reproductive Technology (ART) evaluated CoTs generated via three distinct strategies: Zero-shot, Random Few-shot (using shallow examples), and Selective Few-shot (using diverse, high-quality examples). These expert ratings were compared against evaluations from a state-of-the-art AI model (GPT-4o).

**Results:** The Selective Few-shot strategy significantly outperformed other strategies across all human evaluation metrics ($p < .001$). Critically, the Random Few-shot strategy offered no significant improvement over the Zero-shot baseline, demonstrating that low-quality examples are as ineffective as no examples. The success of the Selective strategy is attributed to two principles: "Gold-Standard Depth" (reasoning quality) and "Representative Diversity" (generalization). Notably, the AI evaluator failed to discern these critical performance differences. The clinical reliability of synthetic CoTs is dictated by strategic prompt curation, not the mere presence of examples.

**Conclusions:** We propose a "Dual Principles" framework as a foundational methodology to generate trustworthy data at scale. This work offers a validated solution to the data bottleneck and confirms the indispensable role of human expertise in evaluating high-stakes clinical AI.

**Keywords**: Chain-of-Thought; Large Language Model; Assisted Reproductive Technology; Explainable AI (XAI); Human-in-the-loop evaluation

# Introduction

Assisted Reproductive Technology (ART) represents a cornerstone of modern medicine, offering solutions for millions facing infertility [1]. The clinical decision-making process in ART is exceptionally complex, requiring the synthesis of high-dimensional patient data, including baseline characteristics and medical history. This process is time-consuming and fraught with risk for both clinicians and patients, as minute variations in treatment protocols can lead to significant adverse outcomes. Furthermore, clinicians must navigate patients' personal values and ethical considerations, demanding a highly personalized and explainable approach to care [2].

Recent advancements in Artificial Intelligence (AI), particularly Large Language Models (LLMs), have demonstrated considerable promise for answering medical questions, addressing clinical case challenges, and augmenting clinical diagnosis [3–7]. Within Clinical Decision Support Systems (CDSS), these technologies can help synthesize large amounts of data, facilitating more comprehensive and standardized therapeutic strategies. However, while general-purpose LLMs like ChatGPT-4 and Gemini are powerful, their training on broad, non-specialized data limits their utility in niche medical domains. Consequently, high-performing clinical AI applications are typically fine-tuned from general models using curated, domain-specific datasets [8–10]. The actual bottleneck, however, is not a lack of raw clinical data, but a profound lack of explainable data—data that records not just what decision was made, but why. This meticulous, expert-level reasoning, often captured as a Chain-of-Thought (CoT), is the very fuel required to train AI models that are not just accurate, but also trustworthy and scalable to clinicians. To move beyond 'black-box' predictions, models require structured reasoning pathways, or CoT data, which simulate clinical logic and enhance explainability [11,12]. The challenge, therefore, narrows down to a scarcity of expert-authored CoT data. The manual creation of such a dataset on a large scale is prohibitively expensive and time-consuming, presenting a significant barrier to progress in explainable medical AI.

To address this challenge, a promising direction is to leverage the generative capabilities of Large Language Models (LLMs) to synthesize clinical Chain-of-Thought (CoT) data at scale. While this offers a scalable solution to the data bottleneck, it hinges on a critical, unverified assumption: the clinical reliability of the generated content. In a high-stakes domain like Assisted Reproductive Technology (ART), this assumption cannot be taken for granted.

Therefore, this study is designed to examine this uncertainty through a rigorous, head-to-head empirical comparison. Figure 1 presents the conceptual framework of our comparative evaluation study. We hypothesize that a Selective Few-shot strategy, meticulously crafted with diverse and deeply reasoned examples, will significantly outperform both a baseline Zero-shot approach and a naive Random Few-shot strategy. To test this, we developed a novel prompting framework and validated it through a blinded evaluation protocol where senior clinicians assessed the quality of CoTs from all three strategies. In a secondary analysis, we further contrast these expert assessments against a state-of-the-art AI evaluator (GPT-4o) to critically examine the current capabilities and limitations of automated evaluation paradigms. Ultimately, this work aims to establish a foundational, evidence-based methodology for the trustworthy generation of clinical reasoning at scale.

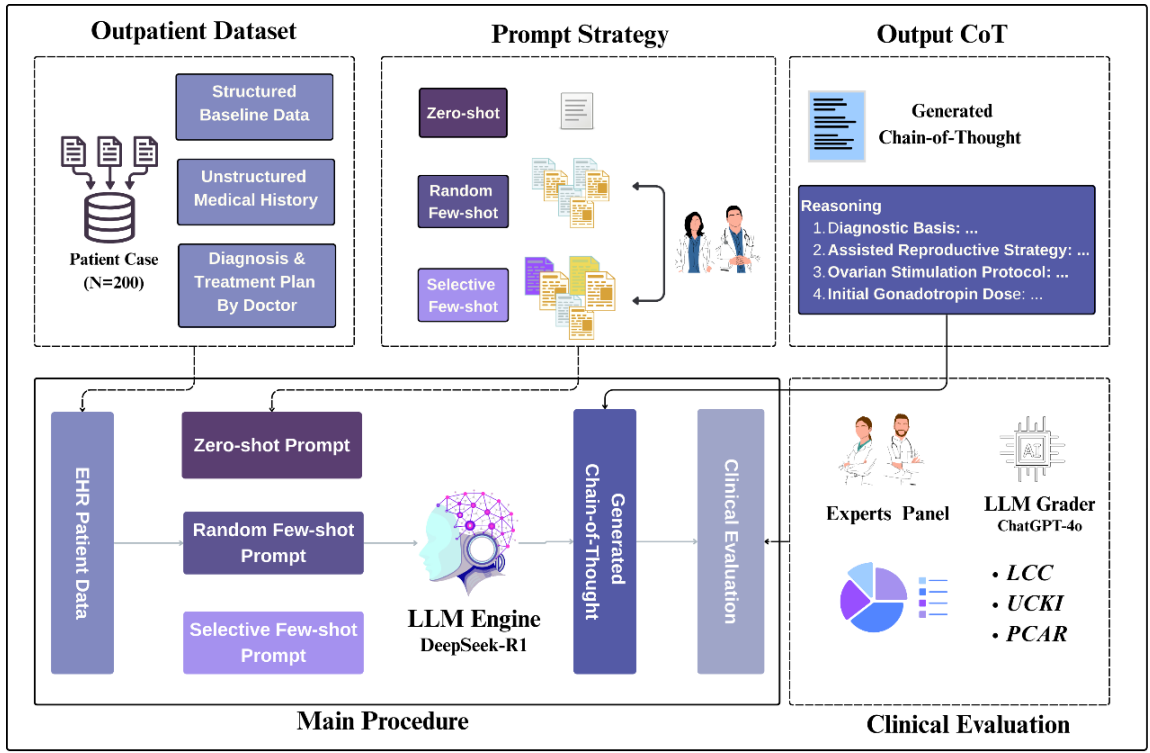

**Fig. 1: Conceptual Framework of the Comparative Evaluation Study.**

The study workflow begins with a standardized patient case (N=200) as the input. Three distinct prompting strategies are evaluated in parallel pipelines: (1) **Zero-Shot**, which uses instructions only; (2) **Random Few-Shot**, which uses five randomly selected examples; and (3) **Selective Few-Shot**, which uses a curated set of 6 diverse examples representing all major ART categories in the cases. Each strategy, powered by the same LLM engine, generates a unique Chain-of-Thought (CoT). All three generated CoTs are then subjected to a rigorous, blinded "Doctor-in-the-Loop" evaluation by two parallel assessors: **human clinical experts** (the gold standard) and a state-of-the-art **AI evaluator (GPT-4o)**. This dual evaluation process yields the final reliability scores and rankings for each strategy.

*LLMs in Healthcare*

Since the launch of ChatGPT-4, Large Language Models have rapidly spread into many industries, such as education, finance, and healthcare. For instance, Google's Med-PaLM 2, a leading specialized healthcare model, achieved 86.5% accuracy on the MedQA benchmark. Furthermore, its responses were preferred over those of generalist physicians in 65% of expert evaluations [13]. The LLMs are now employed in many healthcare-related workflows, ranging from medical documentation assistance to clinical differential diagnosis [5,14–16]. However, to effectively address highly specialized tasks, these models are typically fine-tuned from pre-trained LLMs using carefully curated datasets. Despite their impressive capabilities, current LLMs usually function as black boxes, producing outputs without offering interpretable reasoning. In clinical practice, however, physicians often require not just answers but also transparent explanations. This requires models beyond black-box behavior and providing interpretable, step-by-step reasoning processes. The increasing reliance on LLMs has also intensified the demand for high-quality data [17]. Alarmingly, some predict that the global supply of novel text data may be exhausted by 2050, and image data by 2060 [18]. In the healthcare domain, the situation is even more critical: clinical data is not only scarce but also highly sensitive and expensive to obtain. As a result, a central challenge emerges—how can we build datasets that are both sufficiently large and clinically trustworthy to support transparent, reliable medical AI systems?

***Synthetic Data in Healthcare***
To overcome the data shortage in healthcare, researchers are increasingly turning to Large Language Models (LLMs) to create synthetic data. This approach is promising for several reasons. It allows for data generation at scale, addressing issues of data scarcity and privacy [19,20]. Furthermore, synthetic data can be tailored to balance underrepresented patient groups, potentially improving model robustness and fairness [21]. Generative models have demonstrated remarkable success in these areas, with some studies showing that LLM-generated narratives can be indistinguishable from those written by physicians [22]. This potential, however, is inextricably linked to a profound challenge: reliability. While LLMs can mimic the style of clinical text, ensuring the factual accuracy and clinical plausibility of the content is a far more difficult task. For instance, models have been used to generate both structured Electronic Health Records (EHRs) and unstructured clinical notes [23], but in both cases, the risk of hallucination—where the model generates incorrect or nonsensical information—poses a significant threat in high-stakes medical applications. Therefore, the core challenge moves beyond mere data generation to a more fundamental question of trust. While studies have shown that synthetic data can be effective for certain human labeling and fine-tuning tasks [24–28], these applications often involve relatively straightforward data points. The problem is magnified when the task requires complex, multi-step medical reasoning. In such scenarios, "synthetic" must not equate to "inaccurate." This underscores the urgent need for rigorous evaluation methods, not just for the data points themselves, but for the underlying reasoning processes that produce clinical decisions. Our work focuses on this critical next step: assessing the reliability of synthetically generated reasoning paths.

*Chain-of-Thought*

Chain of thought (CoT) is a prompt engineering technique that enhances the output of LLMs, particularly for complex tasks involving multi-step reasoning. It facilitates problem-solving by guiding the model through a step-by-step reasoning process by using a coherent series of logical steps[29]. This approach has been shown to significantly elevate performance on a wide range of complex reasoning tasks in general domains, especially for arithmetic problems and logical reasoning tasks[30,31]. To enhance the reasoning ability in domain-specific tasks, researchers have started fine-tuning the models with CoTs[12]. Within the medical domain, the potential of CoT is particularly compelling. Its step-by-step nature aligns naturally with the differential diagnosis and clinical reasoning processes employed by physicians. Consequently, researchers have begun to apply CoT prompting to improve accuracy on medical question-answering benchmarks and in practice diagnosis[11,32]. More importantly, CoT offers a crucial pathway towards explainable AI (XAI) in medicine. By externalizing the model's reasoning processes, CoT allows clinicians to scrutinize, understand, and ultimately trust the AI's recommendations, which is a prerequisite for its safe integration into clinical workflows.

The application of CoT is rapidly evolving. Beyond simple prompting, a new frontier in clinical AI is the 1) fine-tuning of models on datasets enriched with Chain-of-Thought data to build inherently more explainable systems, which, however, immediately confronts the fundamental bottleneck of medical AI; and the 2) prohibitive cost and time required for expert clinicians to manually author thousands of high-quality reasoning paths for a training set. An intuitive and scalable solution is to leverage foundational LLMs to synthetically generate these CoTs, creating a cost-effective pathway to train the next generation of trustworthy medical models. However, this entire paradigm hinges on a critical, yet largely unexamined, question: 3) Is the reliability of synthetically generated Chains of Thought (CoTs) adequate to support their application in complex clinical scenarios? Literature to date offers little guidance. Most research focuses on the extrinsic value of CoT (i.e., improving final answer accuracy), with scant attention paid to the intrinsic reliability of the reasoning itself. A model fine-tuned on flawed, albeit synthetically generated, logic could learn to produce seemingly correct answers for the wrong reasons—a risk that is unacceptable in clinical practice. Furthermore, standardized, expert-driven protocols for assessing the clinical validity, coherence, and faithfulness of machine-generated reasoning are conspicuously absent. Our study is designed to directly fill this foundational gap. Before the field can confidently use synthetic CoT for model training at scale, we must first have a rigorous method to measure its reliability. Therefore, we propose and implement a blinded, expert-led evaluation framework to answer the fundamental question: how reliable is synthetically generated sophisticated clinical reasoning, and what is the best prompting strategy to elicit it from LLMs?

## Methods

### *Data source*

A set of selected masked EHRs from West China Second University Hospital was considered in this study, and the study was approved by the Ethics Committee of West China Second University Hospital of Sichuan University (ID:2022288). The EHRs have been manually reviewed and corrected to ensure data accuracy. These cases were recorded during 2020-2022 in the Infertility outpatient department, and all the personal information was masked for privacy protection. From the manually reviewed dataset, we randomly selected 200 cases as our evaluation set, covering a variety of Assisted Reproductive Technologies (ART). These ARTs are broadly categorized into three generations: In Vitro Fertilization (IVF), Intracytoplasmic Sperm Injection (ICSI), and Preimplantation Genetic Testing (PGT). Each generation includes several clinical subtypes, such as short-protocol IVF and IVF with donor sperm. Among the 200 evaluated cases, IVF accounts for the largest proportion (n = 140, 70%), including standard IVF (n = 116, 58%), IVF with donor sperm (n = 9, 4.5%), and Short-Protocol IVF (Short-time insemination, n = 15, 7.5%). The second most common is ICSI (n = 38, 19%), comprising standard ICSI (n = 26, 13%), IVF+ICSI (n = 5, 2.5%), and TESA + ICSI (n = 7, 3.5%). PGT represents 11% of the dataset (n = 22), including PGT-A (n = 6, 3%), PGT-M (n = 3, 1.5%), and PGT-SR (n = 13, 6.5%).

The dataset consisted of three main components: (1) a structured set of Baseline and Demographic variables, (2) the Preliminary Diagnosis and Treatment Plan, and (3) an unstructured narrative description of the Present Illness History shown in Table 1. The structured baseline data served as the quantitative and categorical foundation for clinical assessment, encompassing key indicators of ovarian reserve such as Anti-Müllerian Hormone (AMH) and baseline Follicle-Stimulating Hormone (FSH) levels. The unstructured narrative provided essential clinical context, offering a detailed account of the patient's medical journey—information critical for nuanced and context-aware medical reasoning. The output data, labeled as Preliminary Diagnosis and Treatment Plan, reflects the clinical conclusions and therapeutic strategies formulated by human experts in prior encounters. This component serves as the ground truth outcome. The LLM's task is to generate a reasoning path that logically connects the patient's input data to this expert-defined outcome. A detailed breakdown of all case data variables, including structured baseline indicators and narrative clinical history, is provided in the Supplementary file. Together, these inputs formed the foundation for CoT generation and model evaluation.

**Table 1: Structure and Description of Input and Output Variables for Each Case.**

| Category | Variable |
| --- | --- |
| **Baseline and Demographics** | Female age |
|  | Menstrual cycle |
|  | Body weight |
|  | Body Mass Index (BMI) |
|  | Anti-Müllerian Hormone (AMH) level |
|  | Duration of infertility |
|  | Gynecological ultrasound findings |
|  | Baseline follicle-stimulating hormone (FSH) level |
| **Present Illness History** | Present Illness History |
|  | Type of infertility |
|  | Controlled ovarian stimulation (COS) protocol |
| **Preliminary Diagnosis and Treatment Plan** | Initial gonadotropin (Gn) dosage |
|  | Preliminary differential diagnosis |
|  | Initial ART strategy |

This table outlines the variables provided to the LLM for each case, categorized into Input (Baseline & Demographic, Present Illness History) and Output (Preliminary Diagnosis and Treatment Plan). These variables form the basis for the CoT generation task.

## *Experiment Design*

To systematically evaluate the reliability of LLM-generated Chain-of-Thought (CoT) and to determine the impact of different prompting strategies, we designed a comparative study. The experiment was structured into three distinct arms, each representing a different level of contextual information provided to the model. Our design philosophy was to create a controlled, stepwise comparison to isolate the effects of in-context examples and the strategy used for their selection.

All three groups utilized the evaluation dataset ( N=200) described in the case records part, making sure of a fair comparison. A capable "Teacher Model" is key to generating better-quality data[33]. Considering the models' performance so far, we used the open source model DeepSeek-R1-671b, which was known for its outstanding reasoning capability, as our consistent model shared by three arms[34]. The core task for the LLM in each arm remained consistent: Generate a detailed, step-by-step clinical reasoning CoT that derived from all the data provided and expert output.

*Group 1: Zero-shot Baseline*

In this group, we aimed to establish a fundamental baseline to evaluate the out-of-the-box clinical reasoning capabilities of general-purpose LLMs when applied to this specialized task. To this end, the model was prompted using a standardized directive instruction, with each clinical case embedded directly into the prompt (see Supplementary file for details). The outputs generated by the model, along with corresponding physician evaluations, served as a performance floor, quantifying the baseline reliability and limitations of an unadapted LLM in handling novel clinical scenarios.

*Group 2: Random Few-shot Prompting*

This experimental arm was designed to establish a baseline for a standard, non-optimized few-shot approach. Its purpose was to measure the impact of providing generic, in-domain examples without a specific selection strategy. For each of the 200 test cases, the prompt was initially prepared with a fixed set of five examples to provide context for the model. These five examples were randomly sampled from our expert-authored data pool, excluding the existing evaluation dataset. The sample set used in the prompt for every test case consisted of four standard IVF cases and one Short-Protocol IVF case, accompanied by a concise reasoning chain authored by domain experts. The prompt structure and instructions were otherwise identical to those in the other arms. A representative example of a few-shot sample, detailing the input data and expert-written CoT, is provided in the supplementary file. This approach represents a 'naive' few-shot implementation. It is designed to test the hypothesis that the mere presence of in-domain examples, even without being specifically tailored to the test case, is sufficient to improve reasoning quality compared to the zero-shot baseline.

For representative qualitative case studies, we report the number of times that the observed patterns or reasoning errors occurred across repeated evaluations. All p values, including non-significant results, are provided exactly in the text or tables. Statistical significance was defined as $p < .05$.

*Group 3: Selective Few-shot Prompting*

This arm represents our proposed method and was designed to test the hypothesis that a deliberately curated set of diverse examples would improve reasoning reliability and generalization. Instead of random sampling, this approach utilized a clinically informed, representative selection strategy. Physicians curated a set of six exemplary cases from a pool of records not included in the N=200 evaluation set (to prevent data leakage). These six examples were specifically chosen to represent the full spectrum of major ART categories present in our dataset, including IVF (standard IVF, Short-Protocol IVF, and IVF with Donor Sperm), ICSI, TESA+ICSI, and PGT (PGT-A). Their reasoning part was carefully crafted and covered all critical steps. The complete prompt is provided in the supplementary file. For every test case, this same curated set of 6 diverse examples was prepended to the prompt. The purpose of this strategy was to provide the model with a comprehensive and representative 'knowledge base' within the prompt itself. We hypothesized that exposing the model to a diverse range of clinical scenarios would enhance its ability to generalize its reasoning, particularly for less common case types, leading to a more robust performance compared to the 'naive' random-sampling approach.

In summary, this three-arm design allows for a multi-faceted analysis of CoT reliability. The comparison between Group 1 and Group 2 will isolate the general benefit of using in-context examples. The critical comparison between Group 2 and Group 3 will determine whether our proposed selective prompting strategy provides a statistically significant improvement over a random baseline. Together, these comparisons will build a clear evidence-based argument for the importance of a well-designed prompting strategy in generating reliable clinical reasoning.

## *Evaluation Metrics*

### *Physician Evaluation*

The evaluation was conducted by a panel of board-certified reproductive physicians. When disagreements arose, another senior physician was included for the final decision. Each evaluator possesses over ten years of clinical experience in the field of ART. Prior to the formal evaluation, a calibration session was held where all evaluators scored ten cases together. Any discrepancies were discussed to ensure a consistent understanding of the criteria.

In this study, we created an evaluation metric involving three dimensions: Logical Coherence and Clarity, Utilization and Coverage of Key Information, and Plausibility and Clinical Accuracy of Reasoning. All generated CoTs were scored by the 5-Likert scale (1=Very Poor, 5=Excellent) across three key dimensions of reliability, as detailed in Table 2.

*AI Grader Evaluation*

In addition to manual evaluation conducted by human experts, we implemented a supplementary evaluation component leveraging a widely used LLM verifier [35], GPT-4o, to explore its feasibility as an automated evaluator of clinical reasoning. This design enables a direct comparison between AI-generated assessments and the human expert gold standard, offering insights into the consistency, reliability, and potential utility of LLMs in clinical education or decision support. To ensure comparability, the evaluation criteria provided to the AI model were identical to those outlined in Table 2, including definitions of accuracy, logical coherence, clinical appropriateness, and completeness.

**Table 2. Rubric for the Evaluation of CoT Reliability.**

| Metric | Definition |
| --- | --- |
| Logical Coherence and Clarity | Assesses whether the reasoning process is internally consistent, logically structured, and expressed clearly and understandably. |
| Utilization and Coverage of Key Information | Evaluates the extent to which the reasoning incorporates and addresses relevant clinical data points presented in the input. |
| Plausibility and Clinical Accuracy of Reasoning | Measures whether the reasoning is clinically sound, aligns with standard medical knowledge, and leads to a reasonable interpretation or decision. Deduct points as appropriate across the four parts in the analysis. |

The table defines the three dimensions—Logical Coherence and Clarity, Utilization and Coverage of Key Information, and Plausibility and Clinical Accuracy of Reasoning—used by both human experts and the AI evaluator to assess the quality of generated CoTs on a 5-point Likert scale.

*Statistical analysis*

All statistical analyses were conducted in Python (pandas, SciPy, and statsmodels). Each case (n = 200) was independently evaluated under three prompting strategies (Zero-shot, Random Few-shot, and Selective Few-shot) across three dimensions: Logical Clarity and Coherence (LCC), Utilization and Coverage of Key Information (UCKI), and Plausibility and Clinical Accuracy of Reasoning (PCAR). Results are reported as mean ± standard deviation (SD), with exact n values indicated in tables. Because the same 200 cases were scored under all three strategies, we used one-way ANOVA to test for differences among strategies within each dimension. When the assumptions of ANOVA were not satisfied, analyses were confirmed with the non-parametric Friedman test. Post-hoc pairwise comparisons were performed using two-tailed paired t-tests with Bonferroni correction to account for multiple testing. Exact *p*-values are reported, and significance was defined as $p < .05$.

## Results

All the results were obtained through the evaluation dataset (n = 200), including several kinds of ART. As mentioned above, three metrics were used for evaluation: LCC (Logical Coherence and Clarity), UCKI (Utilization and Coverage of Key Information), and PCAR (Plausibility and Clinical Accuracy of Reasoning). The evaluation was done by a panel of experienced practitioners.

### *General Performance*

Table 3 presents the average scores of each prompting strategy on LCC, UCKI, and PCAR. The Selective Few-shot strategy outperformed both Zero-shot and Random Few-shot approaches across all three metrics. Specifically, it achieved mean scores of 4.56, 4.66, and 4.18, which were significantly higher than those of the Zero-shot strategy (4.18, 4.30, 3.85, respectively; all p < .001) and the Random Few-shot strategy (4.31, 4.42, 3.91, respectively; all p < .001).

Notably, there was no statistically significant difference between the Zero-shot and Random Few-shot groups on PCAR, though samples did improve the model's capability on LCC and UCKI statistically significantly. This suggests that merely incorporating a small number of randomly selected examples—without regard to their clinical relevance or representativeness—may offer limited additional benefit over zero-shot prompting.

**Table 3: The table compares the performance of the "Zero-shot," "Random Few-shot," and "Selective Few-shot" strategies.**

| Strategy | LCC | UCKI | PCAR |
| --- | --- | --- | --- |
| **Zero-shot** | 4.18 (0.56) | 4.30 (0.63) | 3.85 (0.53) |
| **Random Few-shot** | 4.31 (0.64) | 4.42 (0.58) | 3.91 (0.63) |
| **Selective Few-shot** | **4.56 (0.50)** | **4.66 (0.53)** | **4.18 (0.56)** |

Values are mean (SD), n = 200 cases per group. The evaluation metrics are Logical Coherence and Clarity (LCC), Utilization and Coverage of Key Information (UCKI), and Plausibility and Clinical Accuracy of Reasoning (PCAR)

## *Subgroup Analysis*

To further dig into the reasons for Selective Few-shot's winning, we did an analysis grouped by ART. Table 4 presents the scores of three ART generations.

In the largest subgroup, IVF (n=140), a key distinction emerged. While the Selective Few-shot strategy significantly outperformed both other groups across all metrics ($p < .001$ for all comparisons), there was no statistically significant difference observed between the Random Few-shot and Zero-shot strategies ($p = .192$). This indicates that for this patient group, providing topically relevant but shallow examples offered limited performance improvement over a zero-shot baseline.

The analysis of the PGT subgroup (n=22) revealed the clearest advantage for prompt diversity. The Selective Few-shot strategy, which was the only prompt containing a PGT example, scored significantly higher than the Random Few-shot strategy across all three metrics: Logical Coherence (LCC, p = .030), Information Utilization (UCKI, p < .001), and Clinical Accuracy (PCAR, p = .030). Consistent with other findings, the Random Few-shot strategy showed no significant improvement over the Zero-shot baseline in this category (LCC, p = .1708; UCKI, p = .2607; PCAR, p = .1348, respectively).

In the ICSI subgroup (n=38), the Selective Few-shot strategy again demonstrated a measurable advantage. It achieved statistically significant improvements over the Zero-shot baseline in two of the three key metrics: LCC (p = .01) and PCAR (p = .048). While the mean score for UCKI was also highest in the Selective group, this specific comparison did not reach the threshold for statistical significance (p = .0595), though the trend was positive.

**Table 4: Subgroup Analysis by ART Category**

| ART | Strategy | LCC | UCKI | PCAR |
|---|---|---|---|---|
| IVF | Zero | 4.20 (0.57) | 4.34 (0.61) | 3.88 (0.53) |
| | Random | 4.29 (0.67) | 4.44 (0.53) | 3.90 (0.65) |
| | Selective | **4.59 (0.49)** | **4.69 (0.51)** | **4.20 (0.55)** |
| ICSI | Zero | 4.16 (0.59) | 4.29 (0.65) | 3.87 (0.53) |
| | Random | 4.37 (0.54) | 4.45 (0.69) | 3.97 (0.59) |
| | Selective | **4.45 (0.50)** | **4.53 (0.60)** | **4.11 (0.56)** |
| | Zero | 4.09 (0.43) | 4.05 (0.72) | 3.64 (0.49) |
| PGT | Random | 4.32 (0.57) | 4.27 (0.70) | 3.86 (0.56) |
| | Selective | **4.59 (0.50)** | **4.68 (0.48)** | **4.18 (0.59)** |

Values in each table are mean (SD). Subgroup analyses were based on the respective case counts (IVF, n = 140; ICSI, n = 38; PGT, n = 22). This table aims to further investigate the performance differences of various prompting strategies across specific clinical scenarios. To achieve this, we categorized the 200 evaluation cases based on their primary type of assisted reproductive technology (ART) — including In Vitro Fertilization (IVF), Intracytoplasmic Sperm Injection (ICSI), and Preimplantation Genetic Testing (PGT) — and conducted a comparative analysis of evaluation outcomes within each group.

## *Case Study*

As shown in Figure 2, to qualitatively illustrate the stark differences in reasoning quality revealed by our quantitative analysis, we selected a representative and complex case involving Preimplantation Genetic Testing for Monogenic Diseases (PGT-M). This case is particularly illustrative as it requires a multi-layered understanding of genetics, ART procedures, and individualized patient factors. The main mistakes are listed in Table 5.

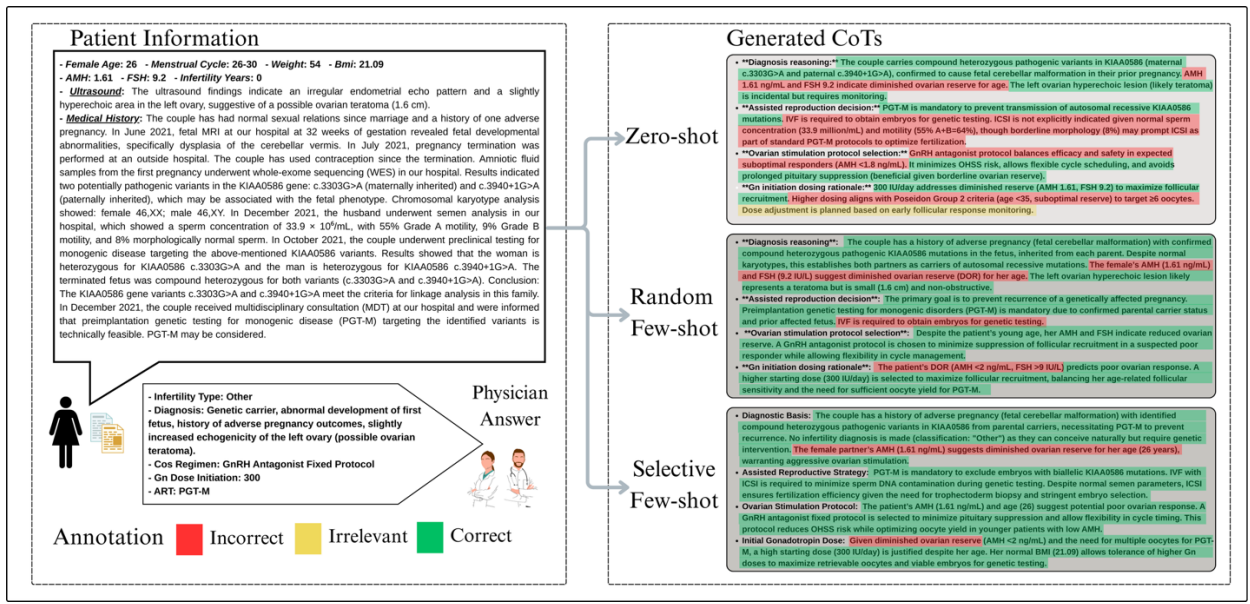

**Fig. 2: Representative PGT-M case illustrating qualitative differences in CoT reasoning across prompting strategies** This figure presents a representative and complex case involving Preimplantation Genetic Testing for Monogenic Disorders (PGT-M), selected to qualitatively illustrate the differences in reasoning quality observed in our quantitative analyses. The left panel shows the patient's clinical information, the correct physician's answer, and the color-coded annotation scheme (red = incorrect reasoning, yellow = irrelevant reasoning, green = correct reasoning). The right panel displays the Chain-of-Thought (CoT) outputs generated under zero-shot, random few-shot, and selective few-shot prompting strategies. Compared with the zero-shot and random few-shot generations, which omitted critical reasoning steps (e.g., the presence of infertility diagnosis, the indication for ICSI, and comprehensive gonadotropin dose considerations), the selective few-shot prompting was more closely aligned with clinical logic and included relevant patient-specific factors.

**Table 5. Common Reasoning Errors in Zero-shot and Random Few-shot CoT Outputs for a PGT-M Case**

| Reasoning Dimension | Flaws in Zero-shot & Random Few-shot |
|---|---|
| Diagnosis reasoning | The model does not mention whether the patient has infertility issues. |
| Assisted reproduction decision | CoT incorrectly assumes that if the male's semen is normal, traditional IVF can be used. |
| Ovarian stimulation protocol selection | The reason for choosing the antagonist protocol in CoT was "greater safety and avoidance of OHSS," without considering the patient's specific circumstances (low AMH, first ovulation induction). |

| **Initial Gonadotropin Dose** | Only AMH levels were considered, without taking into account weight, BMI, or PGT goals (requiring more embryos). |
|---|---|

In this PGT-M case, both partners are carriers of a pathogenic variant in the KIAA0586 gene. During a previous pregnancy, the fetus was found to have a homozygous mutation in KIAA0586, resulting in abnormal brain development and subsequent pregnancy termination. Since then, the couple has been using contraception and therefore does not meet the criteria for an infertility diagnosis. This implies they are still capable of conceiving naturally. Given the autosomal recessive inheritance pattern, there remains a possibility of achieving a normal or carrier embryo through natural conception. However, neither the zero-shot nor the random few-shot prompted CoTs generations mentioned the presence or absence of an infertility diagnosis, which appeared in the selective few-shot prompted CoT.

To avoid the recurrence of a fetus with a homozygous mutation in KIAA0586, preimplantation genetic testing for monogenic disorders (PGT-M) is recommended. Due to the technical requirements of PGT, embryos must be obtained via ICSI (intracytoplasmic sperm injection) to avoid DNA contamination during genetic analysis. While the zero-shot and random few-shot prompted CoTs correctly reasoned the indication for PGT-M, they incorrectly concluded that ICSI was unnecessary because the male partner had normal semen parameters and suggested using conventional IVF instead—an error in clinical reasoning.

In selecting the ovarian stimulation protocol, clinical reasoning typically begins with evaluating the patient's ovarian responsiveness and any prior stimulation history. Although the patient is 26 years old, her AMH level is only 1.61 ng/mL, suggesting a potential for diminished ovarian response. As this is her first controlled ovarian hyperstimulation cycle, a GnRH antagonist protocol was chosen for its controllability and to avoid excessive pituitary suppression. Among the two few-shot prompted CoTs, the reasoning was more aligned with clinical thinking, while the zero-shot CoT emphasized the safety profile of the antagonist protocol (e.g., avoiding OHSS) without clearly reflecting clinical logic.

Regarding the initial gonadotropin dose, factors beyond ovarian responsiveness must be considered. Since this case involves PGT, it is important to optimize the number of oocytes retrieved. Additional considerations include the patient's weight and BMI, as these affect drug sensitivity. However, the zero-shot CoT mentioned only ovarian responsiveness, lacking a comprehensive rationale.

### *Feasibility Analysis of an AI Evaluator*

In stark contrast to the nuanced ratings provided by human experts, the evaluation conducted by the AI agent (GPT-4o) revealed a pronounced ceiling effect. As detailed in Table 6, the mean scores for all three prompting strategies were tightly clustered in a narrow and high-scoring range, between 3.96 and 4.00, suggesting the model perceived all generated outputs as being of similarly high quality.

 Inferential statistical analysis corroborated this observation. A series of one-way ANOVA tests found no statistically significant differences among the three groups for LCC ($F=1.00, p = .37$), PCAR ($F= .79, p = .46$), or UCKI ($F = 2.63, p = .07$). While a post-hoc pairwise t-test identified a marginal statistical difference between the Random Few-shot and Zero-shot groups on the Information Utilization dimension ($t = 2.01, p = .045$), this isolated finding merits cautious interpretation, particularly as the overall ANOVA for this dimension did not reach statistical significance. Collectively, these results indicate that while the AI evaluator could identify generally competent reasoning, it lacked the fine-grained discriminatory power to reliably distinguish the qualitative differences between prompting strategies that were apparent to the human clinical experts.

**Table 6: AI-driven Evaluation of CoT Reliability Across Different Prompting Strategies.**

| Group | LCC | PCAR | UCKI |
|---|---|---|---|
| Random Few-shot | 4.00 (0.00) | 3.98 (0.14) | 4.00 (0.00) |
| Selective Few-shot | 4.00 (0.00) | 3.98 (0.16) | 4.00 (0.07) |
| Zero-shot | 4.00 (0.07) | 3.96 (0.20) | 3.98 (0.14) |

The table presents the mean scores (SD) assigned by the GPT-4o evaluator. Note the high scores and minimal variation across all groups, which indicates a significant ceiling effect in the AI's evaluation.

## Discussion

This study provides a critical evaluation of the reliability of clinical Chain-of-Thought (CoT) reasoning generated by Large Language Models (LLMs), yielding a clear and cautionary conclusion: merely applying standard, uncurated prompting methods is insufficient to ensure clinical reliability. Our physician panel assessments revealed that CoTs generated via both Zero-shot and Random Few-shot strategies consistently scored low on clinical accuracy, with some containing significant reasoning errors. Critically, we discovered that providing topically relevant but shallow examples (Random-shot) offered no tangible improvement in reasoning quality over providing no examples at all (Zero-shot). However, our findings also illuminate a clear pathway toward achieving reliability. We demonstrated that a Selective Few-shot strategy, engineered around a "Dual Principles" framework, significantly enhances the generation of trustworthy CoTs. This framework consists of: 1) Representative Diversity, which endows the model with the ability to generalize across varied clinical scenarios. and 2) Gold-Standard Depth, ensuring that each exemplar reflects expert-level reasoning quality. Crucially, the process of identifying these reliability gaps and validating our solution was only possible through rigorous human evaluation, as our work simultaneously exposed the failure of automated AI evaluators to discern these vital quality differences. Taken together, these findings establish not only a benchmark for assessing clinical reliability but also a foundational methodology for generating trustworthy synthetic data in high-stakes medical AI.

The principle of Representative Diversity was clearly validated in the PGT and ICSI subgroups. The findings provide empirical support for our initial hypothesis. The PGT category shows significantly higher scores, prompted by the selective few-shot approach, which includes an example of PGT-A treatment. The case study also shows errors in understanding and judgment in doctors' viewing, where zero-shot or random few-shot are more likely to make intrinsic mistakes. Notably, in the ICSI category, although the inter-group differences did not reach statistical significance when compared to the random few-shot group, we observed the same trend as in the PGT category—Selective prompting consistently achieved the highest average scores and was significantly higher than zero-shot prompting, which had no difference with the random one. The analyses of both subgroups collectively suggest that a demonstration set covering key procedural subtypes within the domain is essential for enabling the model to evolve from a "specialist" to a "generalist."

Simultaneously, the principle of Gold-Standard Depth was powerfully illustrated in the IVF subgroup. In our main results, we show that the quality of examples may influence the quality of generation. In subgroup analysis, we found that there is no significant difference between the Zero-shot prompting and the Random Few-shot prompting on any subgroup, especially in the IVF subgroup, even if the random arm's sample cases indeed included four standard IVF and one Short Protocol IVF. It performed ineffective learning under this situation. In this case, the reason may be attributed to the reasoning quality in the prompt. In the experiment design section, we mentioned that the random cases have a relatively concise chain of thought. This indicates that the LLM exhibits a strong tendency toward pattern imitation when engaging in in-context learning. A low-quality example tends to elicit correspondingly poor reasoning outputs, even if the model has huge potential in text generation. Therefore, we propose the second core principle for generating high-quality Chains of Thought (CoTs): Gold-Standard Depth. This principle emphasizes that each few-shot example must serve as an expert-level exemplar—logically rigorous, richly detailed, and representative of domain-specific reasoning at the highest standard.

Our findings align closely with a well-established principle in the broader AI research community: data quality often outweighs data quantity[36]. Our work provides strong empirical support for the application of this principle in the high-stakes, domain-intensive context of clinical Chain-of-Thought (CoT) generation. More importantly, we go beyond simply affirming the importance of data reliability—we offer a concrete characterization of what high-quality examples mean in this setting, through our proposed dual principles of Gold-Standard Depth and Representative Diversity. Together, these insights contribute a practical and domain-grounded methodology for realizing data-centric AI in the medical domain.

Another key finding of this study is the limitation of current state-of-the-art large language models (LLMs) when used as evaluators for clinical reasoning. While our human expert assessments revealed substantial differences in reasoning quality across the three prompting strategies—Selective Few-shot, Random Few-shot, and Zero-shot—the scores assigned by the AI evaluator (GPT-4o) showed no statistically significant differences between them. This finding highlights a critical limitation of current LLM-based evaluators in detecting subtle yet clinically meaningful variations in reasoning depth, logical rigor, and contextual accuracy. Although GPT-4o is capable of fluent language generation and general content scoring, it appears insufficiently sensitive to the nuanced features that distinguish high-quality clinical Chains of Thought (CoTs). This 'ceiling effect' serves as a critical warning: in high-stakes medical applications where patient safety is on the line, relying solely on automated evaluation for quality assurance is inherently risky. It reaffirms that domain expert oversight is not merely a "gold standard" for evaluation—it is an essential safeguard that cannot be replaced.

The primary contribution of this study is twofold: establishing a benchmark for how to evaluate and providing a methodology for how to generate trustworthy clinical AI. First, we establish a rigorous, domain-grounded benchmark for evaluating synthetic clinical reasoning. Amid the rapid growth of AI in healthcare, we demonstrate that ensuring clinical validity requires moving beyond automated metrics. Our findings expose the critical limitations of state-of-the-art AI evaluators (e.g., GPT-4o) in detecting subtle but clinically vital reasoning flaws. This "ceiling effect" serves as a critical warning and highlights the indispensable role of structured, blind expert review as an essential safeguard in any high-stakes medical AI development. Second, building on this evaluation framework, we offer a practical solution to the 'explainability data bottleneck'. Through systematic comparisons, we show that a Selective Few-shot prompting strategy—based on the 'Dual Principles' of Gold-Standard Depth and Representative Diversity—substantially improves the quality and reliability of generated CoTs. This offers a feasible, cost-effective blueprint for generating trustworthy synthetic data at scale, without requiring immense annotated datasets.

In summary, our dual contributions—how to evaluate and how to generate—lay a solid foundation for the next generation of Trustworthy Clinical AI. Only when data is both high-quality at the source and rigorously evaluated can we develop AI tools that clinicians can trust and safely integrate into real-world practice.

Although we attempted to determine the liability of AI-generated CoT in complex clinical cases, our cases are currently limited to Obstetrics or Infertility Treatment. To enhance generalizability and robustness, future research should include a more diverse set of complex clinical reasoning cases across different medical departments. Also, the study is based solely on DeepSeek-R1 and could be verified on other models in the future. Our dataset contains 200 diverse cases, but for some subtypes, the number of cases may be too small for statistical analysis. This also reflects a lack of sufficient data for certain types of treatment plans. Further studies might emphasize the importance of solving the long-tail problem in clinical data. Moreover, given the limited context window of LLMs, users may encounter a trade-off when selecting few-shot examples, particularly in domains characterized by substantial subtype diversity. Balancing breadth and depth in example selection becomes a critical challenge under such constraints. Some dynamic prompting techniques were being studied to help achieve a better performance-efficiency trade-off in two practical settings where computational resources or the required performance are constrained[37].

## Data Availability

The datasets generated during and/or analyzed during this study are not publicly available due to institutional case privacy and a large number of interaction dialogs, but are available from the corresponding author on reasonable request. The authors will make the Author Accepted Manuscript (AAM) version available under a CC BY public copyright license.

## Acknowledgements

Part of this study was supported by the Science and Technology Department of Sichuan Province Project (2024YFFK0365); Part of this study was supported by the Natural Science Foundation of Sichuan, China (2025NSFSC1985); Part of this study was supported by the 1·3·5 project for disciplines of excellence, West China Hospital, Sichuan University (ZYYC21004).

## Ethics declarations

Competing interests

All authors declare no conflict of competing interest.

## Supplementary Information

Please see the attached supplementary file.